\documentclass[letterpaper, 10pt, conference]{ieeeconf}      

\IEEEoverridecommandlockouts                              

\usepackage{cite}
\usepackage{amsmath,amssymb,amsfonts}
\usepackage{algorithmic}
\usepackage{graphicx}
\usepackage{comment}
\usepackage[dvipsnames]{xcolor}

\title{\LARGE \bf Meta  Preference Learning for Fast User Adaptation in Human-Supervisory Multi-Robot Deployments
}

\author{Chao Huang$^{1}$, Wenhao Luo$^{2}$ and Rui Liu$^{1*}$
\thanks{$^{1}$ is with the Cognitive Robotics and AI Lab (CRAI), College of Aeronautics and Engineering, Kent State University, Kent, OH 44240, USA. $^{2}$ is with the Robotics Institute, Carnegie Mellon University, Pittsburgh PA 15213, USA. $^{*}$ Rui Liu is the corresponding author, email: ruiliu.robotics@gmail.com.}%
}

\begin{document}

\maketitle
\thispagestyle{empty}
\pagestyle{empty}

\begin{abstract}
As multi-robot systems (MRS) are widely used in various tasks such as natural disaster response and social security, people enthusiastically expect an MRS to be ubiquitous that a general user without heavy training can easily operate. However, humans have various preferences on balancing between task performance and safety, imposing different requirements onto MRS control. Failing to comply with preferences makes people feel difficult in operation and decreases human willingness of using an MRS. Therefore, to improve social acceptance as well as performance, there is an urgent need to adjust MRS behaviors according to human preferences before triggering human corrections, which increases cognitive load. In this paper, a novel Meta Preference Learning (\textit{\textbf{MPL}}) method was developed to enable an MRS to fast adapt to user preferences. \textit{\textbf{MPL}} based on meta learning mechanism can  quickly assess human preferences from limited instructions; then, a neural network based preference model adjusts MRS behaviors for preference adaption. To validate method effectiveness, a task scenario "An MRS searches victims in a earthquake disaster site" was designed; 20 human users were involved to identify preferences as \textit{\{"aggressive", "medium", "reserved"\}}; based on user guidance and domain knowledge, about 20,000 preferences were simulated to cover different operations related to \textit{\{"task quality", "task progress", "robot safety"\}}. The effectiveness of \textit{\textbf{MPL}} in preference adaption was validated by the reduced duration and frequency of human interventions.
\end{abstract}


\section{INTRODUCTION}
An multi-robot system (MRS), where multiple robots with various functions are coordinated to perform tasks, are widely used for complex and large-scale missions, such as disaster search and rescue \cite{8319200,2937074}, site surveillance \cite{roldan2019bringing, scherer2020multi}, and social security \cite{garapati2017game, utkin2017siamese}. Advanced by recent developments in artificial intelligence algorithms, sensor and power technologies, the capability of MRS in task planning, environment perceiving, following human control has been largely improved, leading to wide real-world applications. 

However, for general users it is still challenging to control an MRS to achieve desired mission performance, impeding the wide MRS implementations. One critical cause is the failure of human adaptation. First, human preference is dynamically associated with robot behaviors, task progress, and environmental risks. It is challenging to estimate human preferences by using only pre-defined motions while ignoring dynamic status of robots and environmental constraints. Second, human users have various preferences, imposing the MRS control with different requirements related to in-process robot behaviors, safety concerns, mission efficiency, etc. For example, an MRS customized by aggressive users who focus more on execution speed may fly at high speed in area inspection which is unacceptable for reserved ones who focus more on safety assurance in victim search \cite{canal2019adapting}.
Third, complying with human preferences needs intensive human monitory and corrections, adding a heavy cognitive load onto a human user, further increasing the operation difficulty and decreasing human willingness to use an MRS \cite{prewett2010managing}.

\begin{figure}[t!]
  \centering
  \includegraphics [width=1.0\columnwidth ]{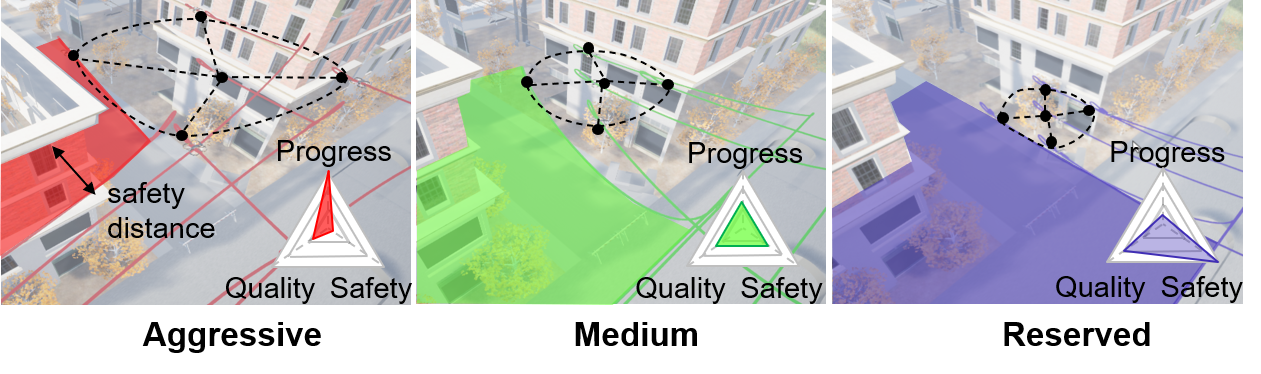}
  \caption{The illustration of using \textit{\textbf{MPL}} to guide MRS adaptation of human preferences, including "aggressive" emphasizing on Progress $>>$ Quality $>$ Safety, "medium" emphasizing on Progress $=$ Quality $=$ Safety, and "reserved" emphasizing on Safety $>>$ Quality $>$ Process. The colored shadow areas denote safety distance to obstacles adjusted according to human preferences. The black dots highlighted UAV locations.}
  \label{idea}
\end{figure}

To adapt to human preferences and finally lower down the operation difficulty of an MRS, in this paper, a novel fast user adaptation model, Meta Preference Learning (\textit{\textbf{MPL}}), was developed to quickly extract preference information from limited user instructions; then integrate it into the robot motion planning model to customize MRS deployments according to user preferences. The method idea is illustrated in a simulated earthquake disaster site, shown in Fig. \ref{idea}, in which \textit{\textbf{MPL}} helps a searching and rescuing MRS team to adapt to different preferences \{"aggressive", "medium", "reserved"\} by balancing the task progress, execution quality, and robot safety (the details are in the Evaluation section). This paper mainly has three contributions:

\begin{enumerate}
\item A novel fast user adaptation method, Meta Preference Learning (\textit{\textbf{MPL}}), is developed to explore control preference from human instructions and adjust MRS motions online to follow human preferences of mission progress, quality and system safety. \textit{\textbf{MPL}} is developed by fusing a novel meta learning into a human preference learning model for quick preference estimation and MRS motion corrections.

\item An active collaboration-calibration manner is designed to improve the human-MRS teaming. Instead of waiting for further human corrections, an MRS actively interprets human instructions to understand performance expectation; then proactively adjust its behaviors to adapt to a human, improving the collaboration quality and meanwhile reducing human cognitive load in robot operations.

\item A novel user-centered cognitive framework is designed by fusing user cognition into robot motion planners. The user cognition refers to general human cognition related to preferences, safety concerns, quality assurance, etc. This framework provides a pipeline to customize an MRS by user cognitive processes, meaningful for future research.

\end{enumerate}

\section{RELATED WORK}

Recently, numerous researches were done to use active learning to learn human preferences. In \cite{sadigh2017active} and \cite{basu2018learning}, users proactively chose more preferred trajectories for an autonomous driving system to learn and behave satisfying human preferences. \cite{daniel2014active} developed a ranking-based learning framework to enable a robot to learn from human-ranked grasping demos. Both of the above approaches represent user preferences based on a linear weighted sum of predefined features such as speed, acceleration and trajectory smoothness. This human preference representation method is inefficient as human preference is highly entangled with task progress, robot performance, and environmental limitations \cite{wilde2020active}. Moreover, these entangled relations are temporally varying, making it unrealistic to define robots' adaptive behaviors for all dynamic moments. To address the problem mentioned above, \cite{christiano2017deep, krening2018interaction, leon2013human} modeled human preference by Markov Decision Process (MDP) by involving reinforcement learning into active learning and learning an adaptation policy guiding robot behaviors. However, an MDP based preference model requires a huge number of interventions to adapt to a new human user. To achieve the goal of faster human preference adaptation, in this work, a novel architecture, meta-learning guided deep neural network, was used to quickly and dynamically update human preference models based on limited human feedback. Most importantly, this fast adaptation method is integrated with an MRS control method, specially designed for an MRS to adapt to the preferences of its human operators.

To reduce users' mental effort, researches investigated query selection strategies in active preference learning for robot performance improvement \cite{biyik2019asking, wilde2020active, racca2019teacher}. \cite{biyik2019asking} utilized the information gain formulation to help a robot to propose easy-answered but informative queries for robots to request guidance while reducing human mental efforts. \cite{wilde2020active} provided trajectories that are easily distinguishable for the user by maximizing the difference of the presented trajectories in the regret aspect. In \cite{racca2019teacher}, an order optimization strategy was developed to propose queries targeting entities that are related to each other, giving users few continuous context switching to reduce the workload of the human teachers, making them reply faster and consequently increasing the human preference learning. However, above approaches mainly focus on query delivery instead of query generation which is more critical for retrieving useful human guidance compared with query delivery. Moreover, the sub-optimal queries generated by robot based on its current understanding of human preference mix unsatisfied and human preferred behaviors together, consequently hindering the human preference learning process. Therefore, in this work, human instructions on MRS performance factors, such as in-process robot behaviors, mission execution quality, and robot system safety were explicitly interpreted as motion adjustments. Explicit human instructions with interactive preference learning can reduce the negative impact of unsatisfied robot behaviors contained in sub-optimal queries, furthermore speed up user preference adaptation.

\section{PRELIMINARIES}
\subsection{Definitions}
Consider an MRS with $n$ robots, the position of each robot $i $ is denoted by $x_i \in R^3$. The volume of robots is denoted as $V \subset R^3$. The workspace is given by $W = R^3 \setminus (O+V+safety_{distance})$, where the volume of obstacles $O$ is dilated by $V$ and human preferred safety distance. The human preference based multi-robot flocking process is parameterized by a set of parameters, $\{{inner}, {height}, {speed}, {safety}\}$, which change based on human preference $H$. ${inner}$ denotes the minimal distance between robots; ${height}$ denotes the flying height of robot system; ${speed}$ represents the flying speed of multi-robot system; and ${safety}$ represents the minimal distance between robots and obstacles.

\subsection{Largest convex region in free space}
To compute the largest convex region in a collision-free workspace, a fast iterative method, IRIS, was designed in \cite{alonso2015multi}. Two recurrent steps are contained in this designed method: given an ellipsoid $E$ and the workspace $W$, find the hyperplanes separating these two sets via quadratic optimization; given the current convex region $P$, find the largest $E$ contained in $P$ via semi-definite program. In this work, we used IRIS to guide MRS behavior planning to increase the robustness of finding feasible human preferred flocking parameters. Besides, the waypoints generated by a global path planner $A^*$ was used to increase the robustness of IRIS in finding feasible $P$\cite{bulitko2006learning}.

\subsection{Human preference-based multi-robot flocking control}
In \cite{vasarhelyi2018optimized}, the author designed a tunable self-propelled flocking model, which guaranteed stable behavior with a large flocking speed in confined spaces. In this work, we adopted the flocking method by involving human preferences. The desired velocity is calculated by:
\begin{equation}
    v_i = v_i^{flock} + v_i^{rep} + v_i^{att} + v_i^{saf} + v_i^{hei} + v_i^{ali}
\end{equation}
where $v_i^{flock}$ is flocking speed towards to goal positions. $v_i^{rep} and v_i^{att}$ is short-range repulsion and long-range attraction terms, respectively. $v_i^{saf}$ is repulsion between robots and obstacles. $v_i^{hei}$ is robot flying height adjustment. $v_i^{ali}$ is velocity alignment. An upper limit is introduced to regulate flying speed to prevent over adjusting:
\begin{equation}
    v_i = \frac{v_i}{\| v_i \|}h_{speed}
\end{equation}
where $h_{speed}$ denotes human preferred MRS flying speed. The above-mentioned adjusting velocity terms have been calculated by a simple half-spring model with a cutoff at the specific value determined by human preference, $H$, under/above which velocity terms start working. For detailed mathematical formulations, please refer to the attachment file.

\section{Meta Preference Learning}
The Meta Preference Learning (\textit{\textbf{MPL}}) method is developed to provide a high-level adaptation strategy to guide the control framework introduced in the above section.

\subsection{User model}
Robot motion behaviors are customized for a specific user by iteratively exploiting relations between robot behaviors and human instructions for motion improvements, expressed below.
\begin{equation}
    Ins = \begin{cases} +1 & if(h_* - R_* > 0.1) \\ 0 & if(|h_* - R_*| \leq 0.1) \\ -1 & otherwise \end{cases}
\end{equation}

\noindent $Ins$ denotes human instructions; $R_*$ represents the current MRS motion status and $*$ represents one of the motion behaviors \{robot inner distance (inner), flying height (height), safety distance to the obstacle (safety), averaged team speed (speed)\}; $+1$ denotes that a human prefers to increase $R_*$, $-1$ denotes to decrease, and $0$ denotes no further motion change needed. This formulation considered one human user has different preferences $H$ on MRS performance as tasks proceed and locations vary.

\begin{figure}[t!]
  \centering
  \includegraphics [width=1.0\columnwidth ]{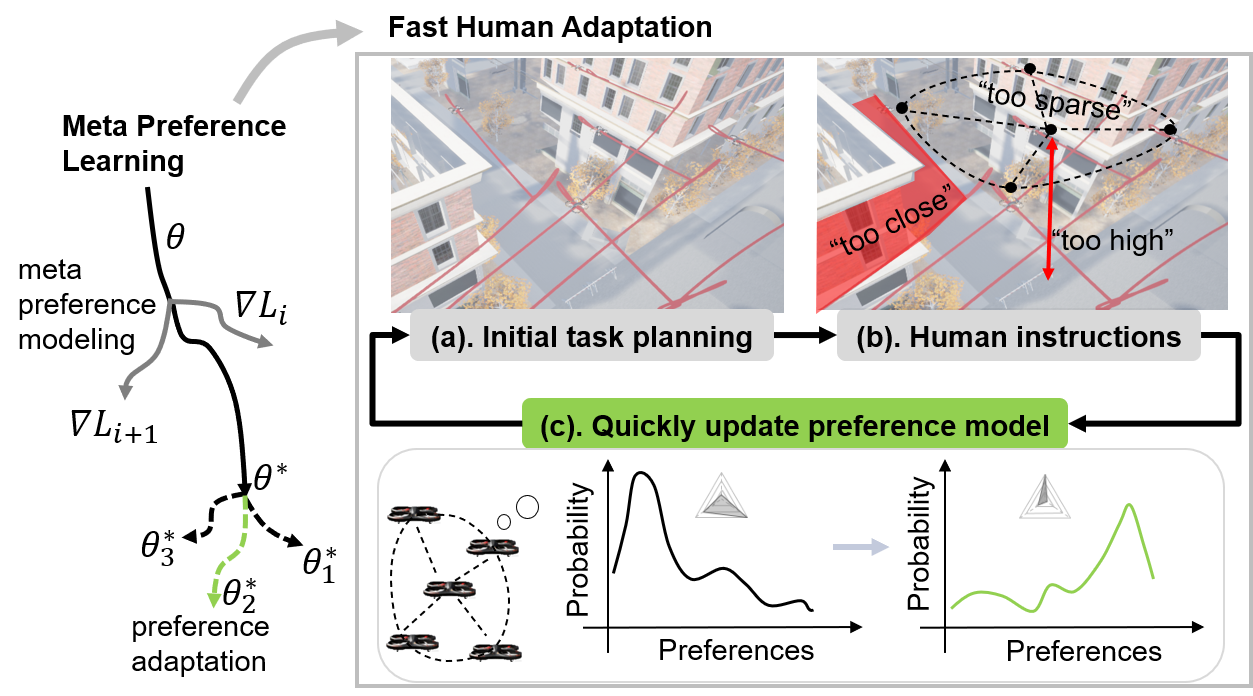}
  \caption{Workflow illustration for using \textit{\textbf{MPL}} for fast human preference adaptation. The left part is the meta-learning framework with a meta learning based preference learning and adaptation. The right part illustrates the fast human adaptation process during MRS deployments. Preferences are adapted by three steps, including initial task planning (a), human instructions for MRS behavior corrections (b), quickly update the deep neural networks supported preference model (c) for MRS motion adjustment (a).}
  \label{method}
\end{figure}

\subsection{Human preference model}
Instead of representing human preference by a linear weighted sum of predesigned features, in this work, a feed-forward deep neural network $M(\theta)$ is used to approximate human preferences according to the human feedback on MRS performance. $\theta$ denotes the parameter set of human preference model. The $M(\theta)$ is effective in exploring the highly nonlinear relation between various preferences, robot behaviors, and task situations.  The input $I$ of $M(\theta)$ includes current robot statuses (positions, speed, accelerations) and environment situations (obstacles position and target position); the output is the human preference value $H$ corresponding to MRS motion status. 

\subsection{Learning framework}
As shown in Fig. \ref{method}, to achieve fast human preference adaptation, we designed a learning framework by using meta learning to guide the neural network based preference learning. Implicit human preferences are extracted from multiple human instructions. During the learning process, current human preference model $M(\theta_{t})$ predicts preference values $\hat{H}_t$ based on current $I_{t}$; then, MRS adjusts motion behaviors by using the preference based multi-robot flocking model designed in Sec. III-C. Then, a user gave instructions $Ins_t$ to further adjust robot behavior. And new human preference values, \begin{math} \hat{H}_{new} = \hat{H}_t + 0.1Ins_t \end{math}, are considered as the training label for time step $t+1$. To evaluate the result of learning $H$, the L2-Norm distance between $H$ and $\hat{H}$ is calculated: \begin{equation}
    L(M(\theta)) = \frac{1}{2}\|H - \hat{H}\|^2
\end{equation}

Learning different users' preferences is considered as as different tasks via meta-learning algorithms, which is fundamentally different from optimizing the human preference model to represent all the users. For each training epoch, a batch of $n = 10$ user preferences, $D_{train}$ is used; each person $p_i$ $(i=1, ..., n)$ gives new instruction as new samples $Ins_i$ to update the training date set  $D_{p_i}^{support}$ and label set $\hat{H}_{new}$, \begin{equation}
    \theta_{p_i}' = \theta - \alpha\bigtriangledown_{\theta}L_{D_{P_i}^{support}}(M(\theta))
\end{equation} where $\alpha$ is the learning rate of inner optimization. With the unused preference samples $D_{p_i}^{query}$, the meta-learning model is trained to maximize performance. The meta objective is defined as following \cite{finn2017model}: \begin{equation}
    min_{\theta} \sum_{D_{p_i} \in D_{train}}L_{D_{p_i}^{query}}(M(\theta_{p_i}'))
\end{equation}
Another meta-learning model "Reptile" in \cite{nichol2018first} is also used to evaluate the reliability of the fast human adaptation method. Instead of applying stochastic gradient descent on the meta-model
parameters $\theta$ to compute the gradient of $L_{D_{p_i}^{query}}(M(\theta_{p_i}'))$, meta-model parameter $\theta$ is directly updated by $\theta_{p_i}'$: \begin{equation}
    \theta \leftarrow \theta + \beta\frac{1}{n}\sum_{i=1}^n(\theta_{p_i}' - \theta)
\end{equation} where the $\beta$ is the meta-learning rate. After meta-training process, an optimal meta preference model $M(\theta^*)$ can be obtained to guide fast human preference adaption with limited human instructions.

\section{Evaluation}
To evaluate \textit{\textbf{MPL}}'s effectiveness in supporting fast human adaptation to an MRS, a multi-robot search and rescue in an earthquake site was designed in a real-gravity simulation environment. The following aspects were validated. (i) the effectiveness of \textit{\textbf{MPL}} in reducing human interventions, which is for robot behavior correction and indicates human cognitive load; (ii) the effectiveness of \textit{\textbf{MPL}} in reducing the intervention duration before MRS achieves qualified performance. In this research, to provide human preferences for model learning and validation, a pioneer user study with about 20 human volunteers was conducted to finally identify three types of user preferences \{"Aggressive", "Medium", and "Reserved"\}. Based on the domain knowledge in MRS deployments and flocking task requirements, motion behaviors of MRS were interpreted as the motion status with value ranges regarding to obstacle and target locations (detailed in Section V.A). Targets in this task were the areas that potentially include victims or has abnormal fire and smoke spots.

\subsection{Experiment setting}

\begin{figure}[t!]
  \centering
  \includegraphics [width=1.0\columnwidth ]{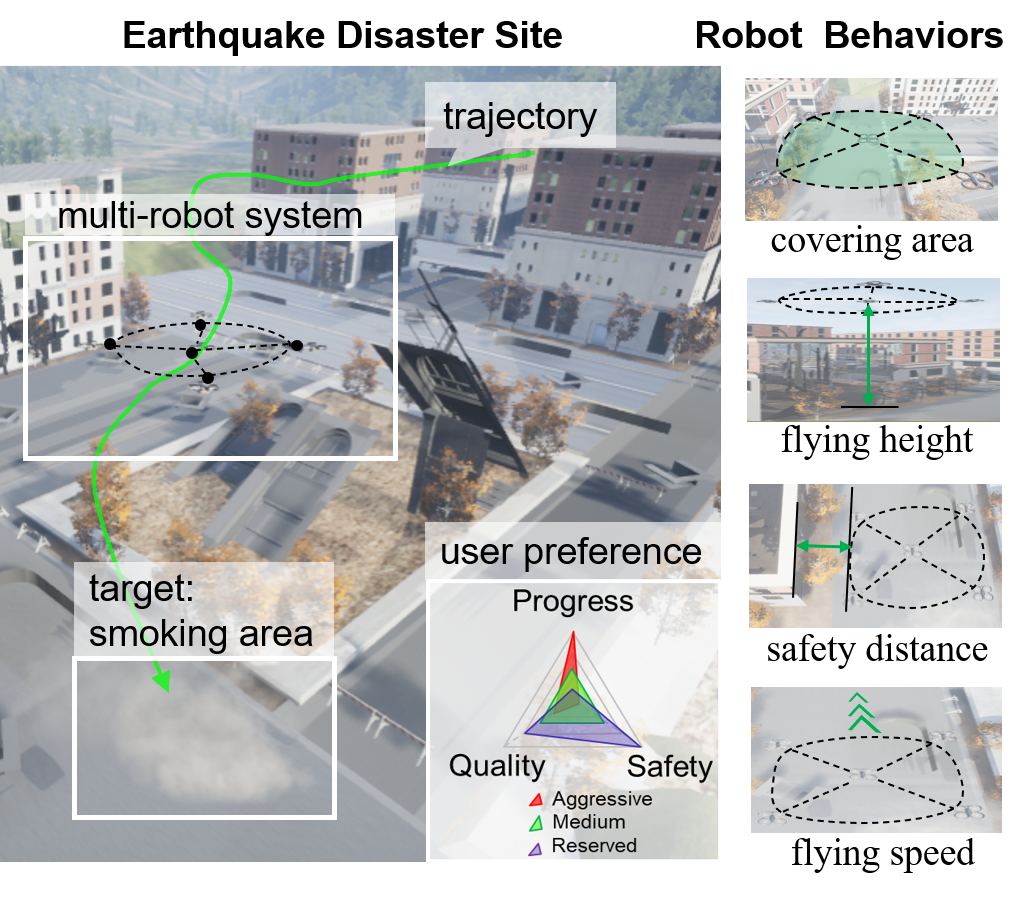}
  \caption{Illustration of experiment setting, include an earthquake disaster site, human preference and multi-robot flocking behaviors. User preferences were identified from pioneer user study regarding to the priority in ensuring \textit{"Progress", "Quality" and "Safety"}. Three typical user preferences, \textit{"Aggressive", "Medium" and "Reserved"}, were used.}
  \label{env}
\end{figure}

Fig. \ref{env} shows the experiment environment "MRS search and rescue in an earthquake disaster site", which was designed by using the open-source platform AirSim \cite{airsim2017fsr} and Unreal Engine Editor \cite{unral2019}. To simulate flocking tasks, the size of the earthquake site was set to 400 $m$ \begin{math} \times \end{math} 400 $m$. The multi-robot team consisted of five UAVs and flocked in the disaster site to inspect abnormal fire and smoke area and identify victims for further assistance. The MRS motion status included four types of flocking behaviors: 1). "covering area" represented by the minimal distance between UAVs was set within [2m, 5m] ; 2). "average flying height" denoted by the average altitude of UAVs was set within [0m, 30m]; 3)."flying speed" represented by the average speed of UAVs was set within [3m/s, 8m/s]; 4). "safety distance to the obstacles" represented by the minimal distance between UAVs and obstacles was set within [0m, 3m].

For searching and rescuing task, three aspects, \textit{"Progress", "Quality" and "Safety"}, were used to illustrate human preferences on MRS deployments. \textit{"Progress"} represents the total mission time needed in a task and is mainly affected by \textit{"covering area" and "flying speed"}; \textit{"Quality"} denotes the in-process searching resolution, in which a higher resolution means a more careful and slower search for potential targets and was affected by \textit{"flying height"} since the higher flying height the less likely for an MRS to discover a target. \textit{"Safety"} was the safety distance of the MRS to obstacles and was affected by \textit{"safety distance"}. As shown in Fig. \ref{env}, three types of human preferences, \textit{"Aggressive", "Medium" and "Reserve"} were involved. \textit{"Aggressive"} was human prefers to finish a task quickly with an more emphasize on mission progress, and less emphasize on mission quality and system safety. \textit{"Reserved"} denotes a human prefers to be cautious about the robot operation by maintaining robot safety and mission quality. The expected robot behavior details corresponding to these three types of human preference are shown in Fig. \ref{beha}. 

To analyze the effectiveness of \textit{\textbf{MPL}} in fast adaptation, the testing process was designed to adapt to multiple humans continually; a baseline method without meta-learning mechanism was used to serve as a control group. Two meta-learning algorithms MAML and Reptile were used to confirm method's reliable performance. The duration of human intervention during one preference adaptation and the number of human interventions needed for achieving qualified performance were compared and analyzed.

\subsection{Result analysis}

\begin{figure}[t!]
  \centering
  \includegraphics [width=1.0\columnwidth ]{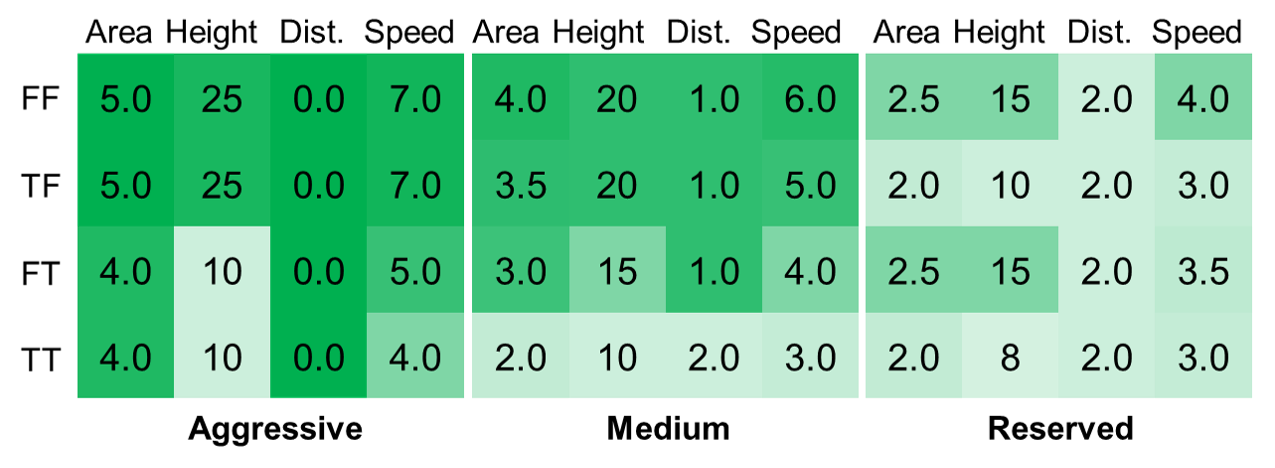}
  \caption{Human preference values for three typical user preferences \textit{"Aggressive", "Medium" and "Reserved"} in four kind of situations ("FF": away from obstacles and target, "TF": near obstacles but away from target, "FT": away from obstacles but near target, and "TT": near obstacles and target). For "Area (covering area)/m", "Height (flying height)/m" and "Speed (flying speed)/(m/s)", darker color represents higher preference value; while for "Dist. (safety distance)/m", darker color represent closer to obstacles.}
  \label{table}
\end{figure}

\begin{figure}[t!]
  \centering
  \includegraphics [width=1.0\columnwidth ]{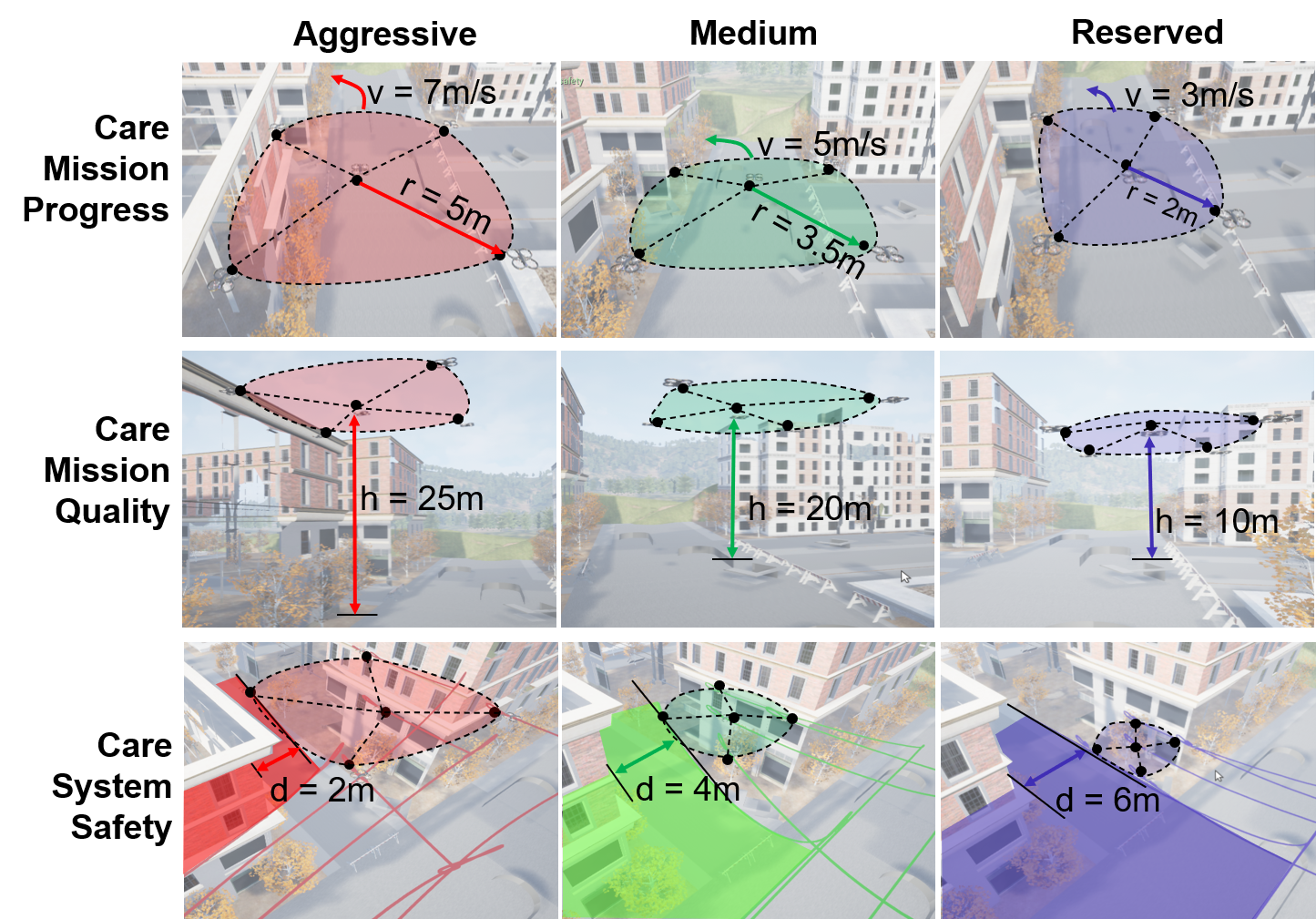}
  \caption{Robot behavior illustration for three typical user preferences \textit{"Aggressive", "Medium" and "Reserved"}. }
  \label{beha}
\end{figure}

\textbf{Human User Study.} With the pioneer user study involving 20 human volunteers, human preferences in MRS operation were identified as three types \{"aggressive, medium, reserved"\}; based on task understanding, each type of preferences was interpreted by volunteers as a value range of a motion behavior regarding to the location of both targets and building obstacles. The details of one sample for \{"aggressive, medium, reserved"\} are listed in Fig. \ref{table}. The sample behavior illustrations for different human preferences are shown in Fig. \ref{beha}. 

\begin{figure}[t!]
  \centering
  \includegraphics [width=1.0\columnwidth ]{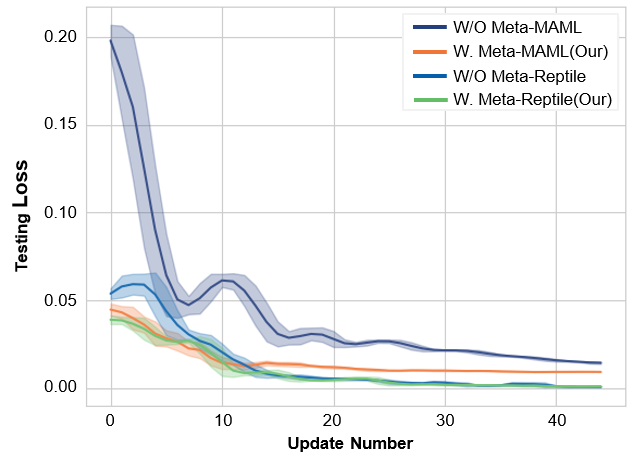}
  \caption{Testing Loss comparison between the developed method \textit{\textbf{MPL}} which have two formats based on two meta learning algorithms \{"MAML", "Reptile"\}, and the baseline methods which are traditional preference models without meta learning support. It shows with the same update number, \textit{\textbf{MPL}} adapted to human preference faster than traditional adaptation methods.}
  \label{metaloss}
\end{figure}

\textbf{Learning Performance Validation.} 
With volunteer guidance, motion behaviors such as \{"covering area", "flying height", "distance to an obstacle", "flying speed" \} within the above ranges were sampled to generate about 20,000 preference samples and each sample denote the operation style of one human; these samples followed into the three categories "aggressive, medium, and reserved"; each category includes about 33\% of the total preference samples. To evaluate \textit{\textbf{MPL}} performance, another 30 human preference samples suggested by human volunteers were used. 
As shown in Fig. \ref{metaloss}, after two times of update, the $L2$ losses between real human preference and predicted human preference of human preference models trained with MAML and Reptile are less than 0.05; while the $L2$ losses obtained from baseline methods are greater than 0.05. It shows the proposed method \textit{\textbf{MPL}} has a faster convergence speed to customize MRS motions for human adaptation.

\begin{figure}[t!]
  \centering
  \includegraphics [width=1.0\columnwidth ]{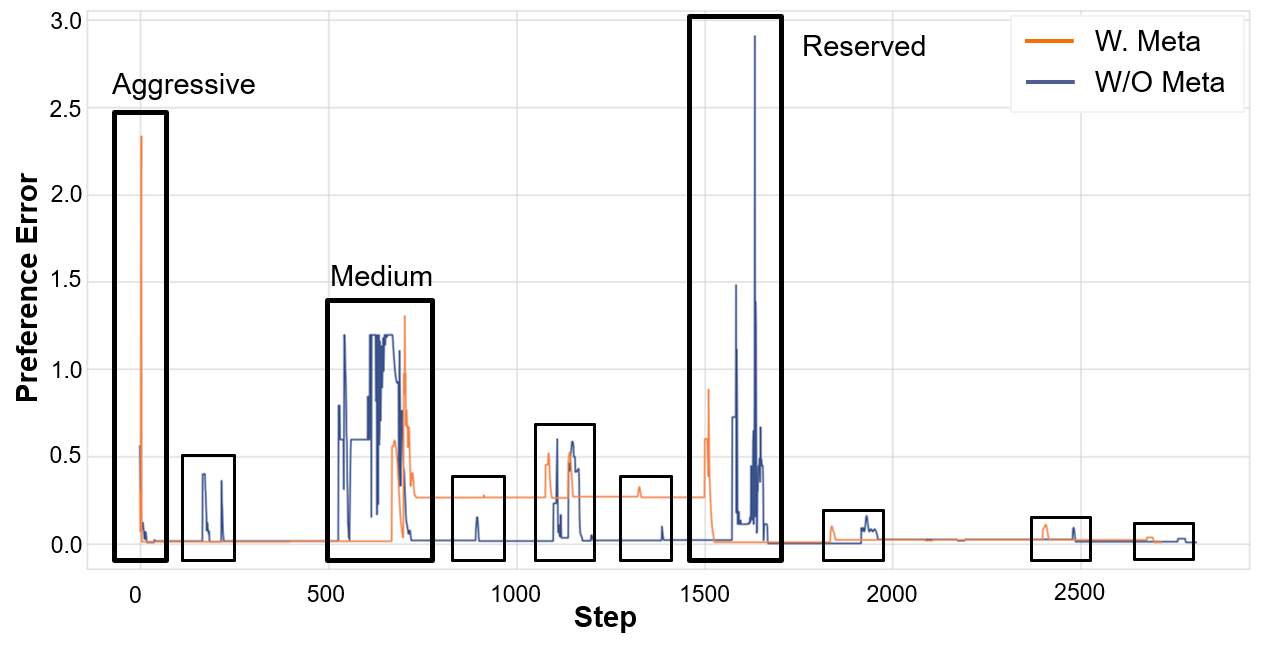}
  \caption{The illustration of the process of multi-robot system continually adapting to three typical user preferences \textit{"Aggressive", "Medium" and "Reserved"}. The black boxes are the human interventions needed for \textit{\textbf{MPL}} model adapting to a specific huam preference.}
  \label{process}
\end{figure}

\begin{figure}[t!]
  \centering
  \includegraphics [width=1.0\columnwidth ]{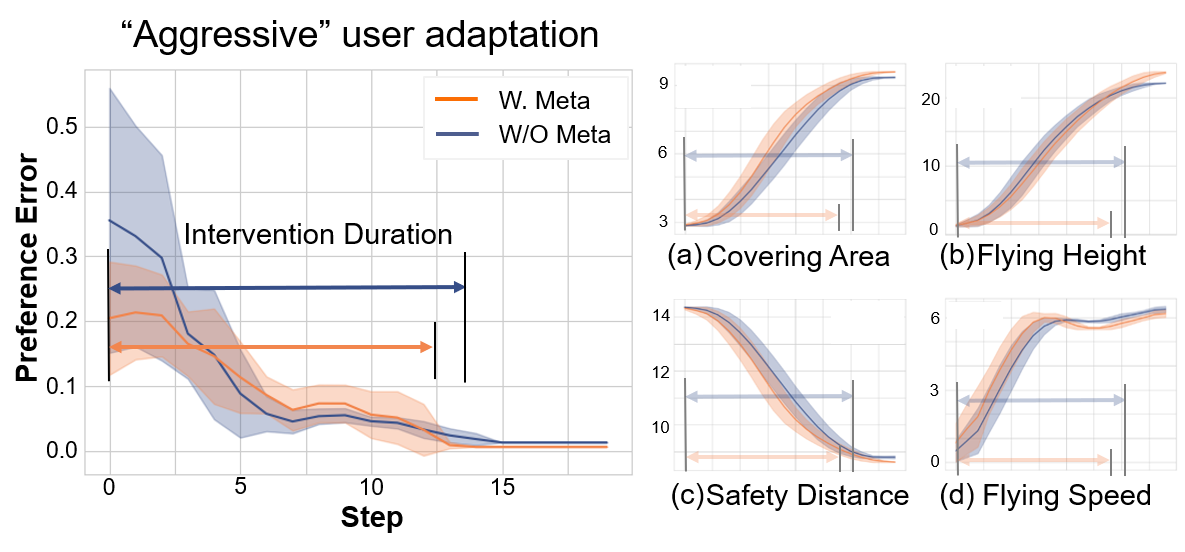}
  \caption{Illustration of robot behaviors adaptation based on explicit human instructions during the first intervention phase of fast adaptation to \textit{"Aggressive"} human preference.}
  \label{behavior}
\end{figure}

\begin{figure}[t!]
  \centering
  \includegraphics [width=1.0\columnwidth ]{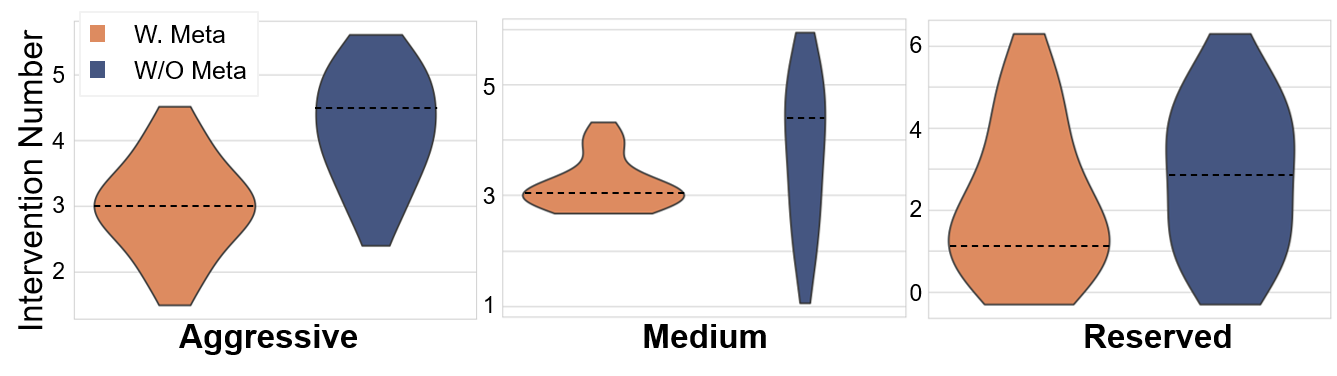}
  \caption{The average number of human intervention phases needed for adapting to three typical user preferences calculated over ten experiments each.}
  \label{num}
\end{figure}

\textbf{Adaptation Effectiveness Validation.} To validate the method's effectiveness in fast preference adaptation, the multi-robot behaviors during one human intervention phase were analyzed; the average number of human interventions needed for the \textit{\textbf{MPL}} to accurately estimate one human preference in MRS motion control was calculated. 
As shown in Fig. \ref{process}, it was a sample process of \textit{\textbf{MPL}} supported human preference adaptation during one complete flocking path. The baseline is the method without meta learning.
The black boxes shown in Fig. \ref{process} were the human interventions needed for \textit{\textbf{MPL}} model adapting to a specific preference. The darker black boxes represent the first human intervention, while the lighter black box denotes the following human interventions following prior interventions. 
To evaluate the effectiveness of \textit{\textbf{MPL}} in fast adapting user preference, the average intervention duration (steps in one box) was calculated. The intervention duration of \textit{\textbf{MPL}} ($12.6 \pm 3.7$) was 42.4\% shorter than that of the baseline method ($21.9 \pm 17$), validating the method's effectiveness in fast adapting to the user preference.
As shown in Fig. \ref{behavior}, one human intervention duration example for the four MRS behaviors was illustrated. As shown in Fig. \ref{behavior}(left), the human intervention duration needed is 2 steps' shorter than that of the baseline method. Fig. \ref{behavior}(a,b,c,d) shows that with the same steps, MRS behaviors in \textit{"covering area", "flying height", "safety distance" and "flying speed"} evolve faster comparing with that guided by the baseline human preference model. Besides, in order to validate \textit{\textbf{MPL}} can fast generalize to similar situations, the average number of human interventions (the number of boxes) needed for an MRS evolving to \textit{"Aggressive", "Medium" and "Reserved"} human preferences was calculated by another 30 human preferences generated by domain knowledge and pioneer user study. As shown in Fig. \ref{num}, \textit{\textbf{MPL}} used fewer interventions than the baseline method for all \textit{"Aggressive", "Medium" and "Reserved"} human preferences. Given intervention frequency and time duration are closely correlated with human cognitive load and operation difficulty, the reduced interventions proved \textit{\textbf{MPL}} positively reduced human cognitive burden and the operation difficulty of a human.

\section{Conclusion and Future Work}
This paper developed a novel fast adaptation method, Meta Preference Learning (\textit{\textbf{MPL}}), to enable multi-robot systems to adjust team motions according to human preferences quickly. To validate method effectiveness, a five-UAV-based multi-robot team was deployed for victim search in an earthquake disaster site; with 20 volunteer based user study, three types of user preferences were identified as \textit{"Aggressive", "Medium" and "Reserved"}. The effectiveness of our proposed method for fast adapting to users was validated by the reduced human intervention frequency, and the interaction times for correcting MRS behaviors. Given the capability of enabling cognitive teaming, \textit{\textbf{MPL}} can be extended to guide flexible teaming multiple robots and even the cooperation between vehicles and human units. In the future, psychology studies could be done to investigate the triggering and maintenance mechanism for some cognitive processes such as safety concern and trust; then, the adaptation method could be enriched and further improved for better human-MRS collaboration.

\addtolength{\textheight}{0cm}   






\begin{thebibliography}{99}


\bibitem{8319200}C. Mouradian, S. Yangui and R. H. Glitho, "Robots as-a-service in cloud computing: search and rescue in large-scale disasters case study," \textit{IEEE CCNC}, pp. 1-7, 2018.

\bibitem{2937074}Z. Beck, Teacy, N. R. Jennings and A. C. Rogers, "Online planning for collaborative search and rescue by heterogeneous robot teams," \textit{Association of Computing Machinery}, 2016.

\bibitem{roldan2019bringing}J. Rold{\'a}n, E. Pe{\~n}a-Tapia, P. Garcia-Aunon, J. Del Cerro, and A. Barrientos, "Bringing adaptive and immersive interfaces to real-world multi-robot scenarios: Application to surveillance and intervention in infrastructures," in \textit{IEEE Access}, pp. 86319-86335, 2019.

\bibitem{scherer2020multi}J. Scherer and B. Rinner, "Multi-robot persistent surveillance with connectivity constraints," in \textit{IEEE Access}, 8, pp. 1509-15109, 2020.


\bibitem{garapati2017game}K. Garapati, J. J. Rold{\'a}n, M. Garz{\'o}n, J. del Cerro, and A. Barrientos, "A game of drones: Game theoretic approaches for multi-robot task allocation in security missions," in \textit{Iberian robotics conference}, pp. 855-866, 2017.

\bibitem{utkin2017siamese}L. V. Utkin, V. S. Zaborovsky, and S. G. Popov, "Siamese neural network for intelligent information security control in multi-robot systems," in \textit{Automatic Control and Computer Sciences}, 8, pp. 881-887, 2017.


\bibitem{canal2019adapting}G. Canal, G. Aleny{\`a}, and C. Torras, "Adapting robot task planning to user preferences: an assistive shoe dressing example," in \textit{Autonomous Robots}, 43(6), pp. 1343-1356, 2019.

\bibitem{prewett2010managing}M. S. Prewett, R. C. Johnson, K. N. Saboe, L. R. Elliott, and M. D. Coovert, "Managing workload in human--robot interaction: A review of empirical studies," in \textit{Computers in Human Behavior}, 26(5), pp. 840-856, 2010.



\bibitem{sadigh2017active}D. Sadigh, A. D. Dragan, S. Sastry, and S. A. Seshia, "Active Preference-Based Learning of Reward Functions," in \textit{Robotics: Science and Systems}, 2017.

\bibitem{biyik2019asking}E. B{\i}y{\i}k, M. Palan, N. C. Landolfi, D. P. Losey, and D. Sadigh, "Asking easy questions: A user-friendly approach to active reward learning," in \textit{arXiv preprint arXiv:1910.04365}, 2019.

\bibitem{basu2018learning}C. Basu, M. Singhal, and A. D. Dragan, "Learning from richer human guidance: Augmenting comparison-based learning with feature queries," in \textit{ACM/IEEE HRI}, pp. 132-140, 2018.

\bibitem{wilde2020active}N. Wilde, D. Kulic, and S. L. Smith, "Active preference learning using maximum regret," in \textit{arXiv preprint arXiv:2005.04067}, 2020.

\bibitem{daniel2014active}C. Daniel, M. Viering, J. Metz, O. Kroemer, and J. Peters, "Active Reward Learning," in \textit{Robotics: Science and systems}, 98, 2014.

\bibitem{christiano2017deep}P. Christiano, J. Leike, T. B. Brown, M. Martic,  S. Legg, and D. Amodei, "Deep reinforcement learning from human preferences," in \textit{arXiv preprint arXiv:1706.03741}, 2017.

\bibitem{krening2018interaction}S. Krening, and K. M. Feigh, "Interaction algorithm effect on human experience with reinforcement learning," in \textit{ACM THRI}, 7(2), pp. 1-22, 2018.

\bibitem{leon2013human}L. A. Le{\'o}n,  A. C. Tenorio, and E. F. Morales, "Human interaction for effective reinforcement learning," in \textit{ECMLPKDD}, 3, 2013.

\bibitem{racca2019teacher}M. Racca, A. Oulasvirta, and V. Kyrki, "Teacher-aware active robot learning," in \textit{ACM/IEEE HRI}, pp. 335-343, 2019.

\bibitem{alonso2015multi}J. Alonso-Mora, S. Baker, and D. Rus, "Multi-robot navigation in formation via sequential convex programming," in \textit{IEEE/RSJ IROS}, pp. 4634-4641, 2015.

\bibitem{bulitko2006learning}V. Bulitko and G. Lee, "Learning in real-time search: A unifying framework," in \textit{Journal of Artificial Intelligence Research}, 25, pp. 119-157, 2006.

\bibitem{vasarhelyi2018optimized}G. V{\'a}s{\'a}rhelyi, C. Vir{\'a}gh, G. Somorjai, T. Nepusz, A. E. Eiben, and T. Vicsek, "Optimized flocking of autonomous drones in confined environments," in \textit{Science Robotics}, 3(20), 2018.

\bibitem{finn2017model}C. Finn, P. Abbeel, and S. Levine, "Model-agnostic meta-learning for fast adaptation of deep networks," in \textit{International Conference on Machine Learning}, pp. 1126-1135, 2017.

\bibitem{nichol2018first}A. Nichol, J. Achiam, and J. Schulman, "On first-order meta-learning algorithms," in \textit{arXiv preprint arXiv:1803.02999}, 2018.

\bibitem{airsim2017fsr}S. Shah, D. Dey, C. Lovett, and A. Kapoor, "AirSim: High-Fidelity Visual and Physical Simulation for Autonomous Vehicles," in \textit{arXiv:1705.05065}, 2017.

\bibitem{unral2019} Epic Games, "Unreal Engine", in \textit{https://www.unrealengine.com}, 2019.


\end{thebibliography}
\end{document}